\documentclass[letterpaper, 10 pt, conference]{ieeeconf}  
\usepackage[T1]{fontenc}
\IEEEoverridecommandlockouts                              

\usepackage{comment}

\usepackage{graphics} 
\usepackage{epsfig} 
\usepackage{amsmath} 
\usepackage{amssymb}  

\title{\LARGE \bf
DMFC-GraspNet: Differentiable Multi-Fingered Robotic Grasp Generation in Cluttered Scenes
}

\author{Philipp Bl\"{a}ttner$^{1,3}$, Johannes Brand$^{2}$, Gerhard Neumann$^{1}$, and Ngo Anh Vien$^{3}$
 \thanks{$^{1}$ Autonomous Learning Robot Lab, KIT}%
 \thanks{$^{2}$ Institute of Production Science, KIT}
 \thanks{$^{3}$ BCAI- Bosch Center of AI}%
}
\begin{document}
\maketitle
\thispagestyle{empty}
\pagestyle{empty}

\begin{abstract}
%
Robotic grasping is a fundamental skill required
for object manipulation in robotics. Multi-fingered robotic
hands, which mimic the structure of the human hand, can
potentially perform complex object manipulation. Nevertheless, current techniques for multi-fingered robotic grasping frequently predict only a single grasp for each inference time,
limiting computational efficiency and their versatility, i.e. unimodal grasp distribution. This paper proposes a differentiable multi-fingered grasp generation network (DMFC-GraspNet)
with three main contributions to address this challenge. Firstly, a novel neural grasp planner is proposed, which predicts a new grasp representation to enable versatile and dense grasp
predictions. Secondly, a scene creation and label mapping method is developed for dense labeling of multi-fingered robotic hands, which allows a dense association of ground truth grasps.
Thirdly, we propose to train DMFC-GraspNet end-to-end using using a forward-backward automatic differentiation approach with both a supervised loss and a differentiable collision loss
and a generalized Q 1 grasp metric loss. The proposed approach is evaluated using the Shadow Dexterous Hand on Mujoco simulation and ablated by different choices of loss functions.
The results demonstrate the effectiveness of the proposed approach in predicting versatile and dense grasps, and in advancing the field of multi-fingered robotic grasping.
\end{abstract}

\section{INTRODUCTION}

Robotic grasping is a fundamental skill required for manipulating objects in cluttered environments \cite{liu2020deep}. Multi-fingered robotic hands, such as the Shadow Hand, mimic the human hand’s structure, enabling complex object manipulations. For instance, the Shadow Hand was used to manipulate a cube in \cite{andrychowicz2020learning}. Data-driven grasp planning for multi-fingered robotic hands has been studied for decades with the aim of finding a hand configuration that provides a stable fixture of the target object inside the hand \cite{bohg2013data}. Grasp planning involves predicting the 6D pose of the robotic gripper, along with determining the joint angles of the fingers for multi-fingered hands. This increases the difficulty by increasing the number of degrees of freedom (DoF). In our study, we use the Shadow Hand, which has 24 degrees of freedom, including 3 positional, 3 rotational, and 18 finger joint angles, to perform robotic grasping. Grasping in a cluttered environment poses additional challenges, as the planned grasps must not collide with other objects in the scene or the table.

Several direct methods have been introduced to predict grasps \cite{liu2020deep,schmidt2018grasping,liu2019generating}. Other methods can directly predict hand configurations, but they often predict only one grasp per prediction or require iterative decoding from different initial conditions to predict multiple grasps \cite{lundell2021ddgc}. Some methods rely on additional components like classical grasp planners \cite{ottenhaus2019visuo}. Contact-GraspNet \cite{sundermeyer2021contact} can predict dense 6D grasp predictions but only for parallel-jaw grippers and is limited to predicting grasps for points outside of the initial object point cloud. In particular, grasp representation with a multi-fingered hand should exploit the object point clouds in a different way from how a parallel-jaw does in Contact-GraspNet. Taking a different approach, these works \cite{liu2020deep,ferrari1992planning} proposing differentiable grasp metrics to take advantage of gradient-based optimization for multi-fingered grasp synthesis. In these work, a differentiable version of the epsilon metric was formulated and used to synthesize grasps. The computation of the epsilon metric is expressed mathematically as a semi-definite programming (SDP) problem. By utilizing the properties of SDP, the gradients of the solution can be derived with respect to the problem parameters, including the gripper pose.

\begin{figure}[t]
      \centering
      \includegraphics[width=0.26\textwidth]{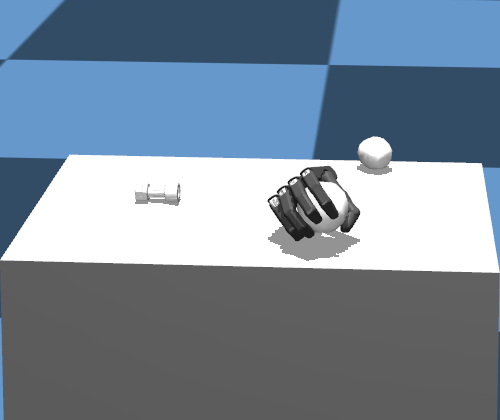}
      \includegraphics[width=0.212\textwidth]{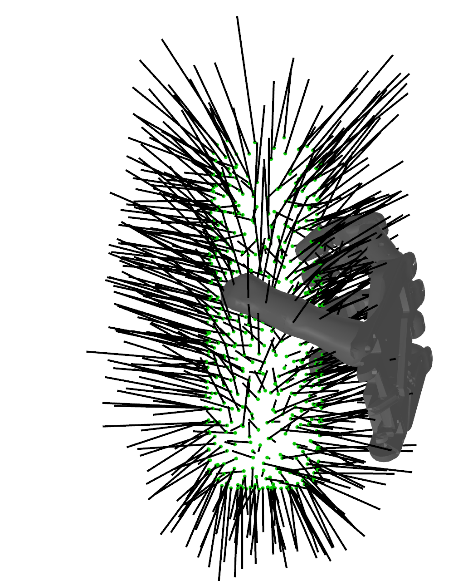}
      \caption{(Left) Grasped object from a multi-object scene with a Shadow Hand. (Right) Our model learns to predict a dense grasp distribution on the input point cloud. Each green point and the connected green line correspond to a prediction of a grasp. Each black line ends in a prediction of the palm translation. The length of the black line is defined by the predicted offset.}
      \label{fig:example_grasp}
   \end{figure}
In this study, we propose a novel approach that combines the strengths of i) Contact-GraspNet by representing and predicting a dense or multi-modal grasp distribution on the input point cloud and ii) the differentiable grasp planning theory that uses differentiable optimization as inductive bias to integrate a generalized Q1 grasp metric and a collision loss in an end-to-end training fashion. The proposed differentiable network extracts visual information by using four different cameras that record the scene from different angles. The proposed approach predicts dense grasp distributions for cluttered scenes and formulates differentiable grasp planning as a continuous optimization. Our method enables the efficient prediction of a diverse set of grasps for a cluttered scene of objects, densely projected to their much less ambiguous contact points, as depicted in Fig. \ref{fig:example_grasp}.


\section{Related Work}

Robotic grasping has been extensively studied for decades \cite{bicchi2000robotic}. Recently, data-driven grasp planning, particularly using deep learning methods, has received significant attention \cite{bohg2013data}. In this section, we review deep learning approaches for grasp planning with multi-fingered robotic hands from visual information.

One popular approach for grasp planning is \emph{direct regression}, where a neural network predicts the grasping posture directly from visual input. The network's input is typically 3D visual information, and it is trained on pre-generated grasp samples for objects from object models. These samples are generated using classical grasp planners, such as \cite{miller2004graspit} and \cite{vahrenkamp2013simox}. In \cite{schmidt2018grasping}, a neural network predicts the palm pose from a single-depth image, and the finger state is determined by closing the hand until it touches the object. Additional post-processing is used to improve the grasp quality. In \cite{liu2019generating}, a model predicts the palm pose as well as the joint angles of the fingers from a 3D occupancy grid. The model is trained on ground truth grasps with an additional collision loss to remove collisions in the predictions with the objects. This concept was further improved in \cite{liu2020deep}, where an additional grasp quality loss was used to improve the quality of the predicted grasp. A neural network is trained to predict a single grasp proposal, and a lower bound of the $Q_1$ metric \cite{ferrari1992planning} is used to fine-tune and improve the predictions of the neural network. However, there work can only predict a single grasp per scene. In contrast, our approach can propose grasps densely on the input point cloud.

Another approach to predict grasps is \emph{probabilistic inference}, also known as grasp success inference. These methods rely on additional grasp planners or sampling algorithms to generate grasp proposals. The neural networks are used to predict a measure that correlates with the grasp success. In \cite{ottenhaus2019visuo}, a model is introduced to predict a grasp quantity measure. The model processes a visual representation and adds additional haptic information. The output of the network then correlates with the success rate. In \cite{lu2020multifingered}, a probabilistic grasp inference model is introduced, building upon earlier work in \cite{lu2020multi}, where a model was trained to predict grasp success. In \cite{lu2020multifingered}, additional models were used to model a grasp distribution. The overall success function is split into conditional priors to improve the overall performance. The predictions are further used to optimize the grasping pose by calculating the gradient of the success metric with respect to the input increase in the grasp success score during evaluation. Though these probabilistic inference based methods can generate multiple grasps per scene, they all require multiple inference rounds. In contrast, our method generates multiple grasps per inference round.

\emph{Generative models}, such as variational autoencoders (VAEs) \cite{kingma2013auto} or generative adversarial networks (GANs) \cite{goodfellow2020generative}, can also be used for grasp planning. These models use generative neural networks to predict grasps from a combination of visual input and random sampling. These methods are able to predict many grasps by using different samples. In \cite{lundell2021multi}, a generative adversarial network is used to predict grasp poses. The proposed grasp poses are further refined by calculating the distance between a shape model of the object to grasp and the initial grasp proposal. The model can be used on scenes of objects by using a segmentation mask to extract the target object. In \cite{wei2022dvgg}, a VAE is introduced to predict grasp poses directly from point clouds. 

\emph{Reinforcement learning} based approach has emerged as a powerful technique that can enable the model to discover
solutions through trial and error. The authors in \cite{abs-1803-09956,abs-1907-11035,FeldmanZVC22} have used RL to learn predictions for both grasp and shift motion primitives for parallel-jaw grippers. The model is trained by grasping simulated objects from a table to maximize the grasp success rate. In \cite{lu2020plafnning}, active learning is employed to generate initial conditions for a neural grasp planner. The method utilizes a multi-armed bandit model to effectively determine whether to optimize an existing sample or generate an additional random start sample. Though these approaches can predict a multi-modal grasp distribution, they often require a large amount of online training data.

\section{Problem Statement}
\label{statement}
We consider the problem of generating dexterous high-DoF robotic hand grasping with physically plausible and non-collision grasp configurations from cluttered scenes consisting of unknown objects. Formally, our method takes as input a recorded point cloud $\cal P$ (possibly, registered from multi-views) and predicts grasps densely projected on the object point cloud. Each grasp is represented as a 6D reference palm point $\mathbf{p}$ (the origin of the palm) and the joint angles of the hand fingers $\mathbf{\theta}$, i.e. $g=\{\mathbf{p},  \mathbf{\theta}\}$. In particular, the Shadow Hand has (6+18)-DoF with: The parameter $\mathbf{p}$ includes a 3-dim translation vector $\mathbf{t}=\{x,y,z\}$ and a 3-dim rotation vector $\{\alpha,\beta,\gamma\}$ of the palm, while parameter $\theta$ includes 18-dim finger joint angles.
\section{Methodology}
Our grasp generation model aims to predict a set of diverse, collision-free grasps for each object in a scene with a single inference from the model. We accomplish this by processing raw visual 3D data and directly predicting the 6D pose of the palm associated with each object point on the input point cloud as well as the joint angles of the fingers. 

\subsection{Grasp representation}
\label{grasp}
We propose an efficient grasp representation that facilitates the acquisition of dexterous grasping skills on complex objects while achieving efficient training. Similar to Contact-GraspNet, we utilize a contact grasp representation, where the distribution of the palm's translation, denoted by $\mathbf{t} \in \Re^3$, of successful ground truth grasps is mapped to their corresponding contact points, denoted by $p_{obj}$. However, Contact-GraspNet restricts visible contact points to lie on surfaces observable with a depth sensor, leading to the representation of their 3D location using nearby points in a recorded point cloud. Nonetheless, using such contact points directly to place the multi-fingered hand can result in an inadequate grasp pose for the fingers.

or multi-fingered hands, the state of the hand is characterized by the pose of the palm and the joint angles of the
fingers. In contrast to Contact-GraspNet, our model is trained to predict these values directly in which the palm’s origin
point is not necessary to be projected directly to points on the input point cloud. To simplify the learning task, we express
the representation of the palm pose differently as follows. First, we use the 6D continuous representation of rotations to describe the orientation of the robot palm, i.e. use two 3-dim vectors $\{\mathbf{a},\mathbf{b}\}$ and reconstruct the orientation matrix through the Gram-Schmidt process proposed in \cite{zhou2019continuity}. The 6D continuous representation can be adversarially transformed into a more conventional representation, such as quaternions. The advantage of this representation is that it eliminates any discontinuities, resulting in more stable training \cite{zhou2019continuity}.
For the expression of the palm position, we use a grasp representation inspired by \cite{morales2006integrated}. Instead of directly calculating it, we utilize an object point, denoted by $p_{obj} \in \Re^3$, to calculate the translation of the palm as follows:
\begin{align}
\mathbf{t}_{palm} = p_{obj} + \mathrm{offset}_{x,y,z} \label{eq:palm_pos}
\end{align}
where $\mathrm{offset}_{x,y,z}\in \Re^3$ is the offset from the reference point to the palm. This approach reduces the prediction interval by only requiring us to predict offsets around the object points. For the joint angles $I \in \Re^{18}$ of the fingers, we directly predict their values. The result of this representation is a 27-dimensional state vector that describes the grasp posture of the Shadow Hand, i.e. $g=\left\{ \mathrm{offset}_{x,y,z}, \{\mathbf{a},\mathbf{b}\}, I \right\}$ associated with each contact point $p_{obj}$.

\subsection{Grasp Data Generation}
We generate different table-top scenes of training and testing. The scenes contain different selections of objects.
We use the collection of object models with their \emph{ground-truth grasp poses} from \cite{liu2020deep}. The selection contains object models from different object databases in the YCB testset \cite{calli2015ycb}, the BigBird dataset \cite{singh2014bigbird}, the data grasping set \cite{kappler2015leveraging}, and the KIT object dataset \cite{kasper2012kit}. We withhold the objects from the YCB \cite{calli2015ycb} set for testing. We further use the generated Ground truth grasps from \cite{liu2020deep}. Liu et al. generated the grasps with GraspIt \cite{miller2004graspit}. The labels contain hundreds of grasps for each object. We follow the scene generation from ACRONYM and Contact-GraspNet \cite{EppnerMF21,sundermeyer2021contact} to generate three different datasets, ranging from simple to complex scenes. 
\begin{itemize}
\item TOur first training set contains 1000 samples of one object scene generated using all 533 non-YCB objects,
with random poses. This leads to additional complexity in our dataset. The position of the objects is selected
randomly to still be on the table. We randomly place the objects on the scene in a random quasi-stable pose
generated with Trimesh \cite{trimesh}. 
\item Our second set of training data contains table-top scenes
of multi objects (3 to 5) using a similar pipeline from
ACRONYM \cite{EppnerMF21}. In particular, we place the objects
iterative in collision-free positions on the table similar
to a single object. We generate random positions until
a collision free position for mesh on the table is found.
For the training, we generate 10.000 different scenes to
train our model containing 3 to 5 different objects.
\end{itemize}

We render depth images of the scenes before the training. The depth images look at the table-top from all for sides of the table. During the training, we extract a point cloud from the depth images. Furthermore, we use the generated poses of the objects from the scene creation to transform the ground truth grasps in a canonical state to objects in the created scene. We match those grasps to object points in the input point cloud to generate dense annotations of the point cloud. We match the grasps by using a reference point on the palm of the hand. We calculate the distance $d_t$ between the ground-truth reference point $p_{p}$ of the palm origin and the object point $i$-th $ p_{o,i}$. In addition, we calculate the signed distance $d_n$ along the normal direction of the mesh object $o$ as follows
\begin{align}
d_t= \|p_{p} -p_{o,i}\|_2, \quad  d_n = n_{o,i} * (p_{p} -p_{o,i})   . \label{eq:distance}    
\end{align}
We match any ground-truth grasp to an object point in the point cloud, if the translation of the palm is closer than $5mm$ to this object point and has a positive distance in the normal direction. Specifically, for object point $i$, there is a set of ground-truth grasps whose reference point $p$ satisfies the matching
\begin{align}
g_i = \{ g_p | d_t <= 5mm \quad \& \quad d_n > 0\}. \label{eq:groundtruth}    
\end{align}
The set of matched grasps $g_i$ is a non-unique matching for each point $i$ per object $o$. An object point with a non-empty set $g_i$ is set with positive label, otherwise negative.

\subsection{Network Architecture}
Our model targets to predict dense grasp proposals for points on the input point cloud, $F: PCL \mapsto {\cal G}$ that maps from input point clouds in $\Re^{N\times 3}$ to a grasp configuration space. As our output grasp distribution is dense, we define ${\cal G} \in \Re^{m \times 27}$, where $m$ is the number of predicted grasp points and $27$ is the dimension of each grasp configuration as defined in \ref{grasp}. In our implementation we predict $m=512$ points. Fig. \ref{fig:example_grasp} (Right) shows an intuitive result of a dense grasp distribution on the input point cloud.

\begin{figure}[thpb]
      \centering
      \includegraphics[width=0.47\textwidth]{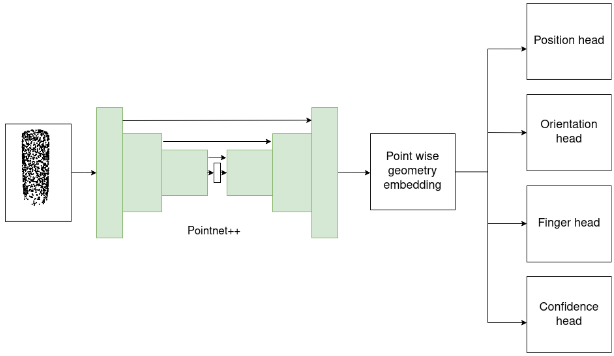}
      \caption{Structure of the backbone network. The pointnet++ \cite{qi2017pointnet++} predicts a point wise geometry embedding for the point cloud. The head networks predict the indidual components.}
      \label{model_backbone}
   \end{figure}

Similar to Contact-GraspNet \cite{sundermeyer2021contact}, we are based on PointNet++ \cite{qi2017pointnet++}, and our network also uses a segmentation-style structure. This allows the feature network to extract an embedding for individual points. We use a U-shaped network as a feature-extracting backbone. The layers of the backbone are based on the set abstraction modules from PointNet++. We employ 4 downsampling modules followed by 3 up-sampling modules. The network input receives a downsampled point cloud of $N=2048$ points. The first downsampling module reduces the number of points to 512. This is the stage that we use for the predictions of the network. The backbone network predicts a 128-dimensional feature vector for each of these 512 points. The network structure is depicted in Fig. \ref{model_backbone}.

We deploy a head network to transform the feature representation for each point to our grasp representation. We split the grasp representation into the position offset, orientation, and the finger joint angles of each grasp prediction. We use one head network for each part. The motivation of using different heads for the 3 different parts is that each part has a different scale and unit. Separating each prediction allows each head network to learn a specific part. The heads are simple 3-layered MLP networks. The first two layers of the head network are linear layers with the relu activation function which also use batch normalization \cite{ioffe2015batch}. The following last linear layer predicts the outputs. The last layers do not have an activation function. The output dimensions of the head network corresponds to the dimensions of the grasp representation.

The network predicts $m=512$ grasps for each scene. We want to be able to select the best grasps from the predictions. For this, we deploy an additional head network to predict a 1-dim confidence value. This is the same head network but with an additional Sigmoid activation function at the output.

We further want to transform a model of the hand to the predicted state in the scene. For this, we calculate the pose of each joint of the hand. Using predicted joint states as input, we apply the forward kinematic layer from \cite{villegas2018neural} to transform the hand mesh model into its corresponding world coordinate state, i.e. all points on the gripper sampled from the hand model mesh are transformed to the predicted pose. Specifically, we use the predicted joint angles to calculate the homogeneous transformation from the parent joint to the each child joints. To enforce the joint angle limits in the predictions, we clamp the joint angles within a predefined range
$
\Theta = \max(\min(\Theta, \Theta_{max}), \Theta_{min})
$. We calculate the position of each link by applying all transformation from the chain to the link. 
This allows us later to calculate loss functions based on collisions of the prediction with the object models and a differentiable $Q_1$ grasp metric.

\subsection{Loss Functions}
We use a task loss function which is a combination of different loss functions which is defined similarly in \cite{liu2020deep}. The task loss is to train our network to predict grasp representations. It can be expressed for a prediction at point $i$ on the input point cloud as follows
\begin{align}
{\cal L}_{task,i} = w_1 {\cal L}_{ch} +  w_2 {\cal L}_{co} + w_3 {\cal L}_{gu} + w_4 {\cal L}_{Q_1^{upper}} \label{eq:task_loss}    
\end{align}
where $w_1,w_2,w_3,w_4$ are weighting coefficients among different loss terms. We now describe each loss term.

\paragraph{Supervised loss} The Chamfer loss ${\cal L}_{ch}$ calculates the distance between the ground truth grasps and the predictions of the network. It was originally introduced in \cite{liu2020deep} to solve to correspondence problem of multi-fingered grasping. The correspondence problem relates to the fact that there are many different possible grasp for each object. Because the model in Liu et. al. \cite{liu2020deep} only predicts a uni-modal grasp distribution, i.e. a single grasp pose per prediction, they implemented this loss in a way that the closest ground truth grasp is selected to the prediction over the entire set of training grasps for this object. However our model can predict a multi-modal grasp distribution, e.g. densely on the point cloud. Therefore we propose to select the closest ground truth from the labels but should also satisfy the matching condition as described in Eq. \ref{eq:groundtruth}. This added condition enables the matching of object points with ground truth grasps locally only. In essence, the Chamfer loss is calculated at each predicted point $i$ with a predicted pose ${\hat g}_i$ as follows
\begin{align}
{\cal L}_{\textrm{ch}, i} = \min_{g_p} \|{\hat g}_i - g_{p}\|^2 \label{eq:chamfer_loss}    \quad {\rm s.t.}\,\, g_p \in G_i
\end{align}
where $G_i$ is the ground-truth grasp pose set at point $i$ as defined in Eq. \ref{eq:groundtruth}. Note that we only calculate ${\cal L}_{\textrm{ch}, i}$ on points with positive label.

\paragraph{Differentiable collision loss} The collision loss is
calculated based on the mesh model of the hands. For this
loss function, we first sample 2000 collision points from
the hand model mesh. We use these points to calculate the
distance between the meshes of the scene and the hand
model. We calculate the distance with respect to every mesh
in the scene including the table, the objects and the links
of the hands (to capture self-collision). Specifically, the
collision loss is calculated as follows
\begin{align*}
{\cal L}_{co} = \frac{1}{2000*L}\sum_{j=1}^{L} \sum_{i=1}^{2000} \max(d^i_j, 0)^2,
\end{align*}
where $d^i_c$ is the signed distance from the closed point of any object mesh to every collision point $i$ and $L$ is the number of meshes in the scene. 

\paragraph{Guidance loss} We further employ a guidance loss to guide the inside of the hand towards the closest mesh. The guidance loss is supposed to minimize the distance between the hand and the surface of the object. The guidance loss uses the distance calculation from the loss functions and will guide the
hand towards the next mesh surface. To calculate the guidance loss, we labeled a set of $V=45$ hand
points, which are located on the inside of the hand. We use these points to calculate the distance between the meshes of the scene and the hand model. We calculate the distance with respect to every mesh in the scene including the table, the objects and the links of the hands. The loss is simply calculated with
\begin{align*}
{\cal L}_{gu} = \sum_j^V ||p_j - p_{\rm{mesh},j}||^2,  
\end{align*}
where $p_j$ is the position of the hand point, and $p_{\rm{mesh},j}$ is the point among all meshes closest to $p_j$. A similar loss function has been introduced in the DDGP
\cite{liu2020deep}. The position of these points is calculated by the collision point layer. The inside
the hand points define a subset of these points. For the multi-object input scenes, all object meshes
are treated as one mesh. The hand is guided towards the closed face of a mesh regardless of the
which mesh the face belongs to.

\paragraph{Differentiable $Q_1$ metric loss} The last term of the loss function is the upper bound of the $Q_1$ metric. The metric was first introduced by \cite{ferrari1992planning}. It describes the quality of a force closure grasp with the minimal wrench to break the grasp. We note that using lower bound of $Q_1$ metric can result higher grasp quality. However its computation is expensive, especially for our model predicting a dense grasp distribution we have to solve multiple SDP problems per forward pass. The upper bound has been first introduced in \cite{schulman2017grasping} and used for training a neural network in \cite{liu2020deep}. We use the notation from \cite{liu2020deep} for the upper bound 
\begin{align}
{Q_1^{\textrm{upper}}} =  \min_ {j=1,...,D}[\max s_j^T M w],  \label{eq:loss_upper}    
\end{align}
where $M$ is a metric tensor to weight the torque components, $w$ is the wrench of the grasp, and $s_j$ is the support of the point. As a result, we define the loss ${\cal L}_{Q_1^{\textrm{upper}}} = e^{-{Q_1^{\textrm{upper}}}}$.

\paragraph{Confidence loss}
In the task loss formulation, the confidence is not regarded in the loss function of the predictions. We adapt the loss augmentation from \cite{wang2019densefusion} to calculate the joint loss function including the confidence with
\begin{align*}
{\cal L}_{\textrm{task, confidence}} = \frac{1}{B}\frac{1}{m} \sum_i^m c_i {\cal L}_{task, i} -w_5 \log(c_i),    \label{eq:conf_loss}
\end{align*}
where $c_i$ is a predicted confidence score at point $i$, $B$ is the batch size and $N$ is number of points in the prediction. We train our network with ${\cal L}_{\textrm{confidence}}$ with $w_5=1$.

\subsection{Implementation Details}
Our model training process employs the Adam optimizer \cite{kingma2014adam} with a carefully chosen learning rate of 0.001 and a learning rate decay of 0.9 after every 10 epochs. The batch size is set to 4. The loss function is evaluated to ensure reliable and robust predictions. Due to memory constraints on the GPU, only the 64 samples with the highest error are selected for backpropagation during training.

Our training methodology involves a two-stage process. In the first stage, we pretrain our models using only the Chamfer loss for 50 epochs to reduce the task loss. This approach serves to guide the network's predictions towards reasonable grasps. The pre-trained model is then fine-tuned with the task loss for an additional 20 epochs, further optimizing the model's performance. The training on the large dataset
of 10.000 scenes take 2 days on a single Nvidia V100 GPU.

To evaluate the effect of different loss functions, we train a model for each combination of loss functions on various training sets. In all cases, we pretrain the models on the data loss to ensure a solid foundation. Our initial loss function uses coefficient $w_1=1, w_2=1, w_3=1, w_4=0, w_5=1$. We fine-tune this model on the first training set of objects with a standard pose to assess the effect of various loss functions. The pre-trained model from the data loss is used as the starting point for fine-tuning different models. 
\section{Experimental Evaluation}
We assess the efficacy of our grasp planner in the simulation environment, Mujoco \cite{todorov2012mujoco}. Our evaluation employs a Shadow Hand model comprising 24 degrees of freedom. To streamline control and isolate the hand, we utilize a hand-only robot model without an attached arm. This approach enables us to evaluate grasp planning performance without
kinematic constraints, further simplifying trajectory planning from the initial hand pose to the target grasp pose. In our
evaluation task, we aim to lift an object from the table and sustain the grip for 2 seconds.

\subsection{Evaluation Data}
We employ the identical scene generation process utilized during training to create test scenes. Our model is
evaluated using 16 selected objects from the YCB test set, \{master\_chef\_can, mustard\_bottle, potted\_meat\_can
banana, bleach\_cleanser, mug, skillet\_lid, wood\_block, mini\_soccer\_ball, softball, baseball, tennis, c\_cups, e\_cups,
f\_cups, i\_cups\}. They are those used in training data generation. To construct 100 distinct test scenes, we randomly
create scenes of i) one object with standard pose, ii) one object with random pose, and iii) randomly spawn 3 to 5
objects and place them on the table. We leverage the same four-camera configuration used during training to capture the
test scenes.

\subsection{Inference and Grasp Execution}

\begin{figure}[t]
      \centering
      \includegraphics[width=0.5\textwidth]{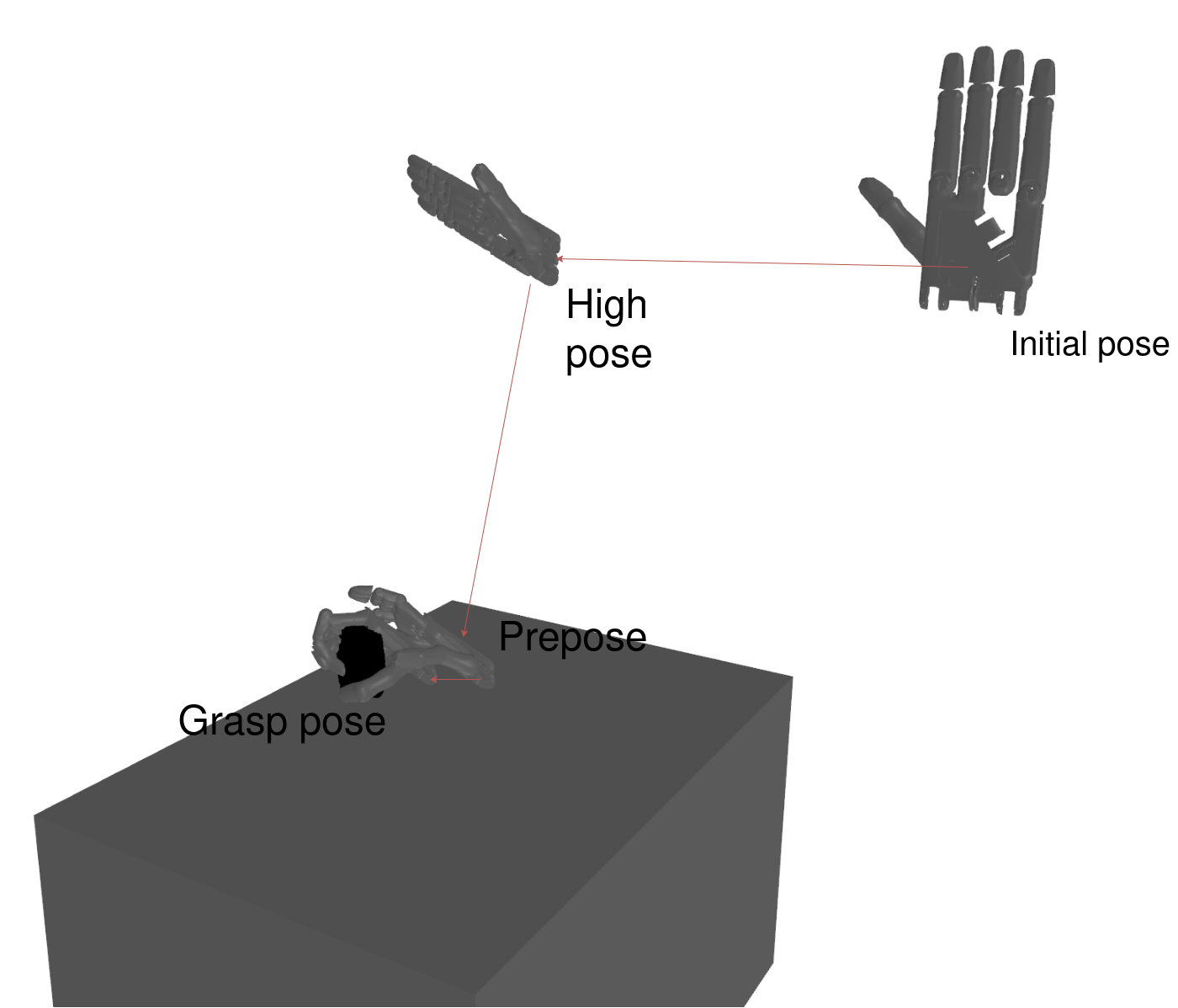}
      \caption{Example of a planned trajectory: The hand is moved from resting pose to the object in four steps. First, the hand is raised to a high pose above the object. Next, it moves to a pre-grasping pose, slightly offset from the final grasp pose along the normal direction. The fingers are open in this pose. Finally, the hand moves into the grasping pose and the fingers close around the object.}
      \label{fig:planned_trajecotory}
   \end{figure}
   
We rendered 4 depth images of the scene using Mujoco
and applied segmentation masks to remove the background,
resulting in a point cloud with 2048 points for all objects in
the scene. We used the same processing sets as in training to
obtain a consistent representation of the global point cloud.
Without any segmentation for specific objects, our trained
model predicts 512 grasp configurations with associated
confidence values. We sort the grasp proposals by confidence
and select the highest-confidence grasps first, pruning those
with confidence below 0.15 to prioritize better grasps.

Once we have selected a target grasp, we start the robotic
hand in a resting pose above the table and move it to the
grasping pose with a set of support hand poses, computed
based on the predicted offset from the pre-grasp pose. To
execute the trajectory, we directly control the position and
orientation of the palm.

To evaluate the results, we lift the object straight upwards to the resting pose height and hold it for 2 seconds. A grasp is successful if the object remains inside the palm at the end of the waiting process.

\subsection{Evaluation metrics}
For the evaluation, we try to select $K = 4$ grasps per
scene to execute. From the list of grasps after filtering using
confidence scores, we select K grasps using the farthest point
sampling algorithm in order to assess the prediction quality
in general.

We evaluate the grasp success with three different metrics.
The first metric calculates the number of valid grasps in the
scenes. The valid grasp rates is the number of successfully
planned grasps from K. The grasp success rate describes
the number of successful grasps out of the successfully
planned grasps. The overall success rate is the combination
out of both. The overall success rate describes the number
of successful grasps out of K evaluated grasps. We evaluate
the success rate of each tested grasps. The reported success
rate is calculated with the number of total evaluated grasps.

\subsection{Simulated Grasping Experiments}
We now report the evaluation results collected on 100
random scenes through the physical execution of predicted
grasps on Mujoco. We report the mean rate and its first
standard deviation.
\paragraph{Evaluation on one object scene of random pose}
Our first experiment evaluates the model trained our first small
training set of simple scene, i.e. single-object scene. We
evaluate for K = 4 grasps for each object. The goal of this
experiment is to evaluate the effectiveness of the different
loss functions. For this, we run our experiment for each of
the model with the same object starting conditions. Before
each evaluated grasp in each scene, we reset the simulation
environment.

The results in Table \ref{tab:result1} show that the pretrained model
with Chamfer loss only can perform poorly but fine-tuning
helps improve either the valid rate or the success rate.
All fine-tuning losses help improve the valid rate, but the
collision and guidance loss reduces the success rate. Fine-
tuning with the Q 1 upper bound loss helps improve the
success rate significantly, however it has the lowest valid rate
improvement. Fine-tuning with all losses including Chamfer
gives the best balanced performance.

\begin{table}[h]
\caption{Evaluation results on one object scene of random pose.}
\label{tab:result1}
\begin{center}
\begin{tabular}{|l|c|c|c|}
\hline
{\bf Model }& {\bf Valid rate} & {\bf Success} & {\bf Overall}\\
\hline
Pre-trained w. ${\cal L}_{ch}$ &0.67$\pm$ 0.02&	0.45 $\pm$0.02& 	0.29 $\pm$0.02\\
\hline
Fine-tuning w. ${\cal L}_{co}$& 		1.0 $\pm$ 0.0& 	0.13 $\pm$ 0.01& 	0.13$\pm$ 0.01 \\ 
\hline
Fine-tuning w. ${\cal L}_{gu}$& 		0.90 $\pm$ 0.02& 	0.35 $\pm$ 0.02& 	0.32 $\pm$ 0.02 \\
\hline
Fine-tuning w. ${\cal L}^{\text{upper}}_{Q^1}$ & 0.75 $\pm$ 0.02 & 	0.73 $\pm$ 0.01 & 	0.55$\pm$ 0.02 \\
\hline
{\bf Fine-tuning w. all}	& 	0.82 $\pm$ 0.02 & 	0.68 $\pm$ 0.02&  	0.56$\pm$ 0.02\\
\hline
\end{tabular}
\end{center}
\end{table}

\paragraph{Evaluation on multi-object scene of random pose}
This experiment evaluates the model trained on our large
dataset of 10.000 complex scenes. The evaluation scenes
contain 3 to 4 objects. We also use 4 grasp trials for each
scene. Due to high computation, we only train and evaluate
the pre-trained model and the fine-tuning with all losses.
The results in Table \ref{tab:result2} show that both models perform well
on both the valid and success rates. Fine-tuning with all
differentiable losses helps improve the overall performance
significantly. These performance is slightly better than the
evaluation on one-object scene because we use a significantly
larger dataset.

The overall results are encouraging and show that the
model is able to predict grasps with improved collision avoid-
ance and grasp quality. In addition, as shown in qualitative
visualization in Fig. \ref{fig:example_grasp}, our model can predict a multi-modal
grasp distribution.

\begin{table}[h]
\caption{Evaluation results on multi-object scene of random pose.}
\label{tab:result2}
\begin{center}
\begin{tabular}{|l|c|c|c|}
\hline
{\bf Model }& {\bf Valid rate} & {\bf Success} & {\bf Overall}\\
\hline
Pre-trained & 0.991 $\pm$ 0.001 & 0.511 $\pm$ 0.001 & 0.507 $\pm$ 0.001\\
\hline
Fine-tuning & 0.994 $\pm$ 0.0 & 0.72 $\pm$ 0.001 &  0.71 $\pm$ 0.001\\
\hline
\end{tabular}
\end{center}
\end{table}

\section{CONCLUSIONS}
In conclusion, we have developed a novel neural network-based approach for grasp planning that can predict a dense, versatile distribution of grasps for a multi-fingered hand given an input of a point cloud. Our most important contribution is to predict a multi-modal grasp distribution, which provides grasp predictions densely on the object point cloud.
In addition, our proposed grasp representation allows for the prediction of multi-fingered grasps on object points, which
significantly improves the effectiveness of the grasp planner. On the other hand, our new loss functions also show that
they help further improve the accuracy and reliability of our model.

While our experiments have shown promising results, there is still much to explore in this direction. 
Future work will include evaluating our approach on different hand models and physical systems to evaluate its performance in
more real-world scenarios. We also plan to integrate the $Q_1$ lower bound to improve the overall fine-tuning performance
of our model. Overall, our fully differentiable grasp planner provides a robust and efficient solution for multi-fingered grasp planning tasks and opens up exciting possibilities for more advanced applications of human-like robotic hands.

\addtolength{\textheight}{-12cm}   




\bibliographystyle{IEEEtran}
\bibliography{my}
\end{document}